\pdfoutput=1
\documentclass[11pt]{article}

\usepackage{booktabs}
\usepackage[normalem]{ulem}
\useunder{\uline}{\ul}{}
\usepackage{amsmath}
\usepackage{tikz}
\usepackage{amsfonts}

\usepackage[]{EMNLP2023}
\usepackage{graphicx}
\usepackage{times}
\usepackage{latexsym}

\usepackage[T1]{fontenc}
\usepackage[utf8]{inputenc}
\usepackage{xspace}
\usepackage{microtype}

\usepackage{inconsolata}
\usepackage{fdsymbol}

\title{Coherent Entity Disambiguation via Modeling Topic and Categorical Dependency}

\author{Zilin Xiao$^{\spadesuit}$ \quad Linjun Shou$^{\clubsuit}$ \quad Xingyao Zhang$^{\clubsuit}$ \\ \bf Jie Wu$^{\clubsuit}$ \quad  Ming Gong$^{\clubsuit}$\thanks{\; Corresponding author.} \quad Jian Pei$^{\vardiamondsuit}$ \quad Daxin Jiang$^{\clubsuit}$ \\
        Rice University$^{\spadesuit}$ \quad Duke University$^{\vardiamondsuit}$ \quad Microsoft STCA$^{\clubsuit}$ \quad \\
        \texttt{zilin@rice.edu} \quad \texttt{j.pei@duke.edu} \\ \texttt{\{lisho, xingyaozhang, jiewu1, migon, djiang\}@microsoft.com}
        }

\newcommand{\ie}{\textit{i}.\textit{e}., }
\newcommand{\eg}{\textit{e}.\textit{g}.\ }
\newcommand{\modelname}{\textsc{CoherentED}\xspace}

\begin{document}
\maketitle
\begin{abstract}
% DONE: adapt Jie's version and new abstract
% Previous Entity Disambiguation (ED) methods use a discriminative paradigm, running candidate entity selection by feeding documents and entities to a length-limited encoder then comparing matching scores between them. 
Previous entity disambiguation (ED) methods adopt a discriminative paradigm, where prediction is made based on matching scores between mention context and candidate entities using length-limited encoders. 
However, these methods often struggle to capture explicit discourse-level dependencies, resulting in incoherent predictions at the abstract level (\eg topic or category).
We propose \modelname, an ED system equipped with novel designs aimed at enhancing the coherence of entity predictions.
Our method first introduces an unsupervised variational autoencoder (VAE) to extract latent topic vectors of context sentences.
This approach not only allows the encoder to handle longer documents more effectively, conserves valuable input space, but also keeps a topic-level coherence.
Additionally, we incorporate an external category memory, enabling the system to retrieve relevant categories for undecided mentions.
By employing step-by-step entity decisions, 
this design facilitates the modeling of entity-entity interactions, thereby maintaining maximum coherence at the category level. 
We achieve new state-of-the-art results on popular ED benchmarks, with an average improvement of 1.3 F1 points. Our model demonstrates particularly outstanding performance on challenging long-text scenarios.

\end{abstract}

\section{Introduction}
% para 1: ed introduction
% para 2: previous bi-encoder paradigm and its flaws
% para 3: current paradigms and their common flaws: length limitation & no explicit coherence constraint -> coherent example 1 (about topic coherence) -> coherent example 2 (about coherence between predicted A and entities to be predicted) -> our motivation
% para 4: our solution discussed
% para 5: our experimental results
% para 6: contributions & summary
Entity disambiguation (ED) is a typical knowledge-intensive task of resolving mentions in a document to their corresponding entities in a knowledge base (KB), \eg Wikipedia. 
This task is of great importance due to its active presence in downstream tasks such as information extraction~\cite{hoffart-etal-2011-robust}, question answering~\cite{yih-etal-2015-semantic} and web search query~\cite{DBLP:conf/wsdm/BlancoOM15}. 

To perform efficient entity disambiguation (ED), one common approach is to encode mentions and candidate entities into different embedding spaces
% , as shown in~\cite{wu-etal-2020-scalable}
. Then a simple vector dot product is used to capture the alignment between mentions and candidate entities. While this method enables quick maximum inner product search (MIPS) over all candidates and efficiently determines the linked answer, it suffers from late and simplistic interaction between mentions and entities~\cite{barba-etal-2022-extend, decao2021}. Recently, researchers have proposed alternative paradigms for solving the ED problem, such as formulating it as a span extraction task~\cite{barba-etal-2022-extend}. In this approach, a Longformer~\cite{DBLP:journals/corr/abs-2004-05150} is fine-tuned to predict the entity answer span within a long sequence consisting of the document and candidate entity identifiers. Another paradigm~\cite{decao2021, decao2022} reduces the ED task to an auto-regressive style in which generation models are trained to produce entity identifiers token-by-token.
% encoder-only style step-by-step decoding!!

Although these approaches offer some mitigation for the late-and-simple interaction problem,
they still exhibit certain vulnerabilities. For instance, Transformer-based encoders impose inherent limitations on input length, preventing the capture of long-range dependency for specific mentions. 
Also, these methods do not explicitly consider coherence constraints, 
% Also, no explicit coherence constraints are posed to these methods, 
while coherence is considered as important as context in early ED works ~\cite{hoffart-etal-2011-robust, chisholm-hachey-2015-entity}. We first propose to condition the model on compressed topic tokens, enabling the system to sustain \textbf{topic coherence} at the document level.
% We argue that 
% instead of using graph methods~\cite{hoffart-etal-2011-robust, DBLP:journals/kbs/Rama-ManeiroVL20} to measure mention-entity coherence, 
% latent topic coherence can guide entity prediction effectively.

In addition, the relationship among entities holds significant importance in the ED task. 
For example, mentions in the document exhibit a high correlation at the category level, where we name it \textbf{category coherence}. However, previous bi-encoder and cross-encoder solutions have overlooked these entity dependencies and focused solely on learning contextualized representations.
Among other works, extractive paradigm ~\cite{barba-etal-2022-extend} neglects entity-entity relation as well; generative EL~\cite{decao2021, decao2022} do possess some dependencies when linking an unknown mention. However, these dependencies arise from the auto-regressive decoding process and require heavy inference compute.

\begin{figure*}[ht]
	\begin{center}
		\includegraphics[width=\linewidth]{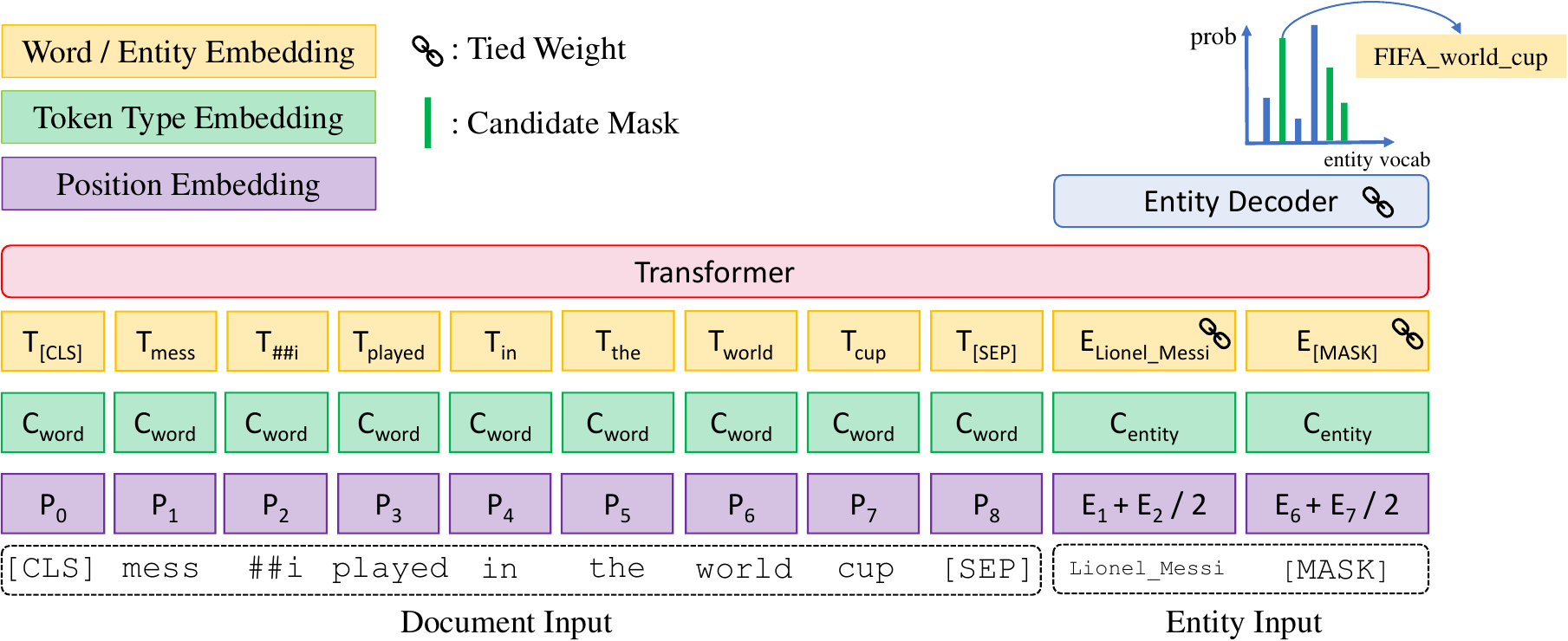}
	\end{center}
	\caption{Overview of our baseline model. 
 Legends are presented in the upper left corner.
 % , where yellow, green and purple blocks denote word/entity, type, and position embedding. 
 Position embedding for an entity is averaged on corresponding document positions, for example, E$_1$ +  E$_2$ / 2 indicates the entity "Lionel\_Messi" is predicted based on document tokens \texttt{mess} and \texttt{\#\#i} at position 1 and 2.}
	\label{baseline_overview}
\end{figure*}

% para 4: our solution discussed
To address the above coherence problem, we propose two orthogonal solutions that target topic coherence and entity coherence, respectively. 
Following previous works that decode masked tokens to link unknown mentions~\cite{yamada-etal-2020-luke, yamada-etal-2022-global}, we present the overview of our coherent entity disambiguation work in Figure \ref{baseline_overview}, where document words and unresolved entities are treated as input tokens of Transformer~\cite{DBLP:conf/nips/VaswaniSPUJGKP17}. 
First, we bring an unsupervised variational auto-encoder (VAE)~\cite{DBLP:journals/corr/KingmaW13} to extract topic embeddings of surrounding sentences, which are later utilized to guide entity prediction. 
% Although generic language encoders such as BERT~\cite{devlin-etal-2019-bert} and auto-regressive language decoders like GPT-2~\cite{radford2019language} are powerful language learners for language modeling and generation, they primarily provide contextualized representations and lack explicit abstract modeling capabilities. 
% The VAE, on the contrary, has demonstrated effective representation abilities in both language modeling and language generation at an abstract level~\cite{li-etal-2020-optimus} (e.g., tense, topic, sentiment). 
By docking these two representative language learners, BERT~\cite{devlin-etal-2019-bert} and GPT-2~\cite{radford2019language}, the variational encoder can produce topic tokens of sentences without training on labeled datasets.
% Together with a vanilla auto-encoder objective (also known as masked language modeling), 
This approach promotes a higher level of coherence in model predictions from an abstract level~\cite{li-etal-2020-optimus} (\eg tense, topic, sentiment).

Moreover, in most KBs, categories serve as valuable sources of knowledge, grouping entities based on similar subjects. To enhance entity-entity coherence from a categorical perspective, we design a novel category memory bank for intermediate entity representations to query dynamically. 
As opposed to retrieving from a frozen memory layer, 
% Instead of interacting with a frozen category embedding layer,
we introduce direct supervision from ground-truth category labels during pre-training. This enables the memory to be learned effectively even from random initialization. 
% Figure \ref{fig:category_memory} illustrates the process of retrieving categories and this process will be elaborated on later.

Named \modelname, experimental results show that our proposed methods surpass previous state-of-the-art peers on six popular ED datasets by 1.3 F1 points on average. Notably, on the challenging CWEB dataset, which has an average document length of 1,700 words, our approach elevates the score from the previous neural-based SOTA of 78.9 to 81.1.
Through model ablations, we verify the effectiveness of the two orthogonal solutions through both quantitative performance evaluation and visualization analysis. These ablations further affirm the superiority of our methods in generating coherent disambiguation predictions. 
% We name our proposed model \modelname and will use this abbreviation in the subsequent sections. 
% Code and checkpoints will be released upon the completion of the review process.

% In summary, our contributions are three-fold as follows:
% \begin{itemize}
%     \item We employ neural approaches to explicitly capture and model both topic and categorical coherence in the ED task.
%     \item We introduce the variational modeling objective and category memory retriever into ED, creating a robust ED system that leverages multi-task learning.
%     \item Through empirical evaluation, we demonstrate that our \modelname surpasses existing state-of-the-art solutions in terms of evaluation metrics. Both visualization and case study support our initial motivations. 
% \end{itemize}

\section{Related Works}
% TODO: Rewrite related works on 06/21, first come up with a summary.

% 1. ED and related works
% 2. Context compressed tokens
% 3. ED as casual language modeling

\textbf{Entity disambiguation (ED)} is a task of determining entities for unknown mention spans within the document. Early ED works~\cite{hoffmann-etal-2011-knowledge, DBLP:conf/i-semantics/DaiberJHM13} commonly rely on matching scores between mention contexts and entities, disregarding the knowledge-intensive nature of ED. Many studies aim to infuse external knowledge into ED. \citet{bunescu-pasca-2006-using} begin to utilize hyperlinks in Wikipedia to supervise disambiguation. \citet{yamada-etal-2020-luke, yamada-etal-2022-global} propose massive pre-training on paired text and entity tokens to implicitly inject knowledge for disambiguation usage. \citet{li2022improving} first leverage knowledge graphs to enhance ED performance. 

Another intriguing feature for entities to differentiate from each other is their type, as entities with similar surface forms (text identifiers) often possess different types. \citet{DBLP:conf/aaai/OnoeD20} disambiguate similar entities solely through a refined entity type system derived from Wikipedia, without using any external information. \citet{ayoola-etal-2022-refined} augment the robustness of such a type system even further. 
Furthermore, many researchers have explored new paradigms for ED. \citet{decao2021} propose using a prefix-constrained dictionary on casual language models to correctly generate entity strings. \citet{barba-etal-2022-extend} recast disambiguation as a task of machine reading comprehension (MRC), where the model selects a predicted entity based on the context fused with candidate identifiers and the document.

The architecture of our system seamlessly blends the advantages of type systems and knowledge pre-training: incorporating type systems as dynamically updating neural blocks within the model, our design enables simultaneous learning through a multi-task learning schema - facilitating topic variation learning, masked disambiguation learning, and knowledge pre-training concurrently.

\textbf{Prompt Compression} is a commonly used technique in language models to economize input space, closely related to topics such as prompt compression~\cite{wingate-etal-2022-prompt} and context distillation~\cite{DBLP:journals/corr/abs-2209-15189}. Their mutual goal is to dynamically generate soft prompt tokens that replace original tokens without hurting downstream application performance. Our topic token design mirrors context compression to some extent but differs in the compression ratio and purpose. 
Our design is more compact; each context sentence gets converted into a single topic token using a variational encoder, and in addition to saving input space, it also retains high-level semantics to guide a more coherent ED.

\section{Methodology}
% In this section, we begin by providing a formal definition of entity disambiguation (ED). Subsequently, we present an overview of our proposed methods. We then delve into the detailed implementations of the topic VAE and category memory, which are essential modules of our model. Last, we address how to build a robust ED system using the multi-task learning objective.
% \subsection{Preliminary}
\subsection{Entity Disambiguation Definition}
Let $ X $ be a document with $ N $ mentions $\{m_1, m_2, \ldots, m_{N}\} $, where each of mentions $m_i$ is associated with a set of entity candidates  $C_i = \{e_{i1}, e_{i2}, \ldots, e_{i|C_i|}\}$. Given a KB with a set of triplets $ G = \{(h, r, t)|h, t \in \mathcal{E}, r \in \mathcal{R}\} $, where $h$, $r$ and $t$ denote the head entity, relation and tail entity respectively, the goal of entity disambiguation is to link mentions $m_i$ to one of the corresponding mention candidates $C_i \subseteq \mathcal{E}$. 

\subsection{Overview} 
% The principle of our proposed methods is illustrated in Figure \ref{baseline_overview}. 
We present the overview of \modelname in Figure~\ref{baseline_overview}.
Following \citet{yamada-etal-2020-luke}, both words in the document and entities are considered input tokens for the BERT model. 
The final input representation sums over the following embeddings:

\textbf{Representation embedding} denotes the topic latent, word embedding or entity embedding for topic inputs, document inputs or entity inputs accordingly. 
% Similar to the baseline setting, 
We set up two separate embedding layers for word and entity input respectively. $ \mathbf{X} \in \mathbb{R}^{V_w \times H}$ denotes the word embedding matrices and $ \mathbf{Y} \in \mathbb{R}^{V_e \times H}$ denotes the entity embedding matrices, where $ V_w $ and $ V_e $ represents the size of word vocabulary and entity vocabulary. 
% Since our evaluation involves a limited number of candidate entities, the entity vocabulary only needs to cover those candidates, resulting in a more manageable entity embedding layer in terms of size and computation. To optimize efficiency, the parameters of the entity embedding layer are tied with the entity decoder at the end of our ED pipeline, following the common practice in text-only pre-training paradigms. Apart from words and entities, we introduce new topic representations for a fixed number of sentences in the document. Refer to Part \ref{topicVAE} for a detailed process of building topic latent.

\textbf{Type embedding} is for discrimination usage. There are three types of tokens available, each of which corresponds to a dedicated group of parameters, $\mathbf{C}_{\text{word}}$, $\mathbf{C}_{\text{entity}}$, $\mathbf{C}_{\text{topic}}$.

\textbf{Position embedding} marks the position of input words and entities, avoiding the permutation-invariant property of the self-attention mechanism. 
% The entity position embedding has also to declare which word tokens are this entity resolving from. This is achieved by applying absolute position embedding for both words and entities. If a mention consists of more than one token, the entity position embedding will average over all corresponding positions.
The entity position embedding also indicates which word tokens the entity corresponds to. This is achieved by applying absolute position embedding to both words and entities. If a mention consists of multiple tokens, the entity position embedding is averaged over all corresponding positions.

% Topic VAE and category memory are introduced below, both of which serve as important auxiliary branches of our proposed method.
% Zilin on 1/27: moving to ED steps

% \end{itemize}

\begin{figure*}[ht]
	\begin{center}
		\includegraphics[width=\linewidth]{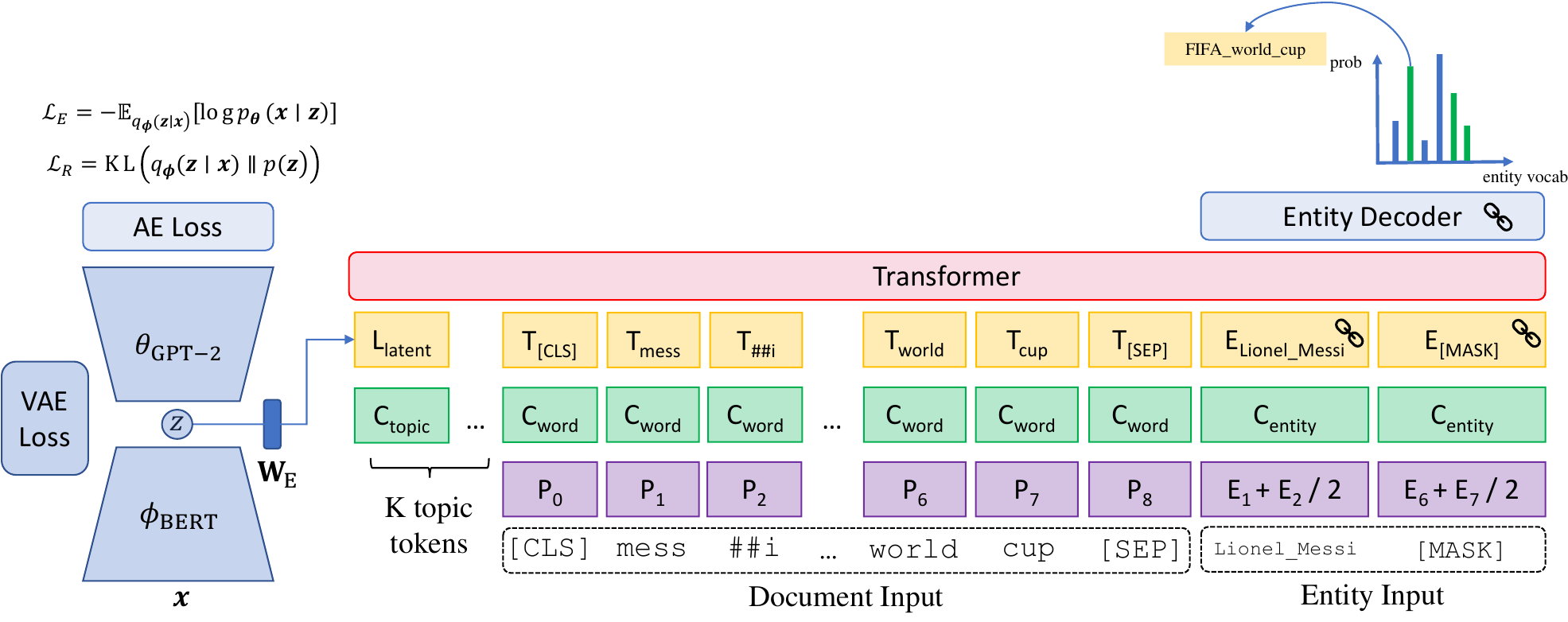}
	\end{center}
	\caption{Topic variational autoencoder injection illustration. $K$ topic sentences are converted into topic tokens appending on the start of model's input. 
 % $\mathcal{L}_E$ and $\mathcal{L}_R$ denote auto
 }
	\label{fig:topicvae}
\end{figure*}

\subsection{Topic Variational Autoencoder\label{topicVAE}}
To preserve maximum topic coherence and optimize input space utilization, we introduce an external component that facilitates topic-level guidance in ED prediction. Among various latent variable models, variational autoencoders (VAEs) have demonstrated success in modeling high-level syntactic latent factors such as style and topic. 
% Our setup is based on such an assumption: a number of sentences will be encoded into a latent vector encoding its topic information after training variational objective on a massive dataset such as Wikipedia corpus. 
Our setup is based on the motivation that, after being trained on a massive corpus using the variational objective, the encoder gains the capacity to encode sentences into topic latent vectors.
The entire topic VAE is composed of two parts, BERT and GPT-2, both of which are powerful Transformer-based language encoder and decoder.

\subsubsection{Encoder\label{vaeEncoder}}
Given a BERT encoder $\text{LM}_\phi$ and input sentence token sequence $x$, we collect the aggregated embedding $ X $ from the last layer hidden states corresponding to the \texttt{[CLS]} token: $ X = \text{BERT}_{\texttt{[CLS]}}(x)$. With this, we can construct a multivariate Gaussian distribution from which the decoder will draw samples. The following formula describes the variational distribution for the approximation of the posterior:

\begin{equation}
\begin{aligned}
& q_\phi\left(z_n \mid \boldsymbol{x}_{\leq n}, \boldsymbol{z}_{<n}\right) = \\
& \mathcal{N}\left(z_n \mid f_{\mu_\phi}\left(x_n\right), f_{\sigma_\phi}\left(x_n\right)\right), \label{eq:encoder_formula}
\end{aligned}
\end{equation}
where $f_{\mu_\phi}$ and $f_{\sigma_\phi}$ denote separate linear layers for mean and variance representations, $\mathcal{N}$ denotes Gaussian distribution and $z$ denotes intermediate information bottleneck. 

\subsubsection{Decoder}
Given a GPT-2 decoder $\text{LM}_\theta$, we first review how to generate a text sequence of length $T$ using such neural models. To generate word tokens of length $ L $, $ \mathbf{x} = [ x_1, x_2, \ldots, x_T ] $, a language decoder utilizes all its parameters $\theta$ to predict the next token conditioned on all previous tokens generated $x_{<t}$, formulated as follows:

\begin{equation}
p(\boldsymbol{x})=\prod_{t=1}^T p_{\boldsymbol{\theta}}\left(x_t \mid x_{<t}\right).
\end{equation}

When training a language generator alone, the decoder is usually learned using the maximum likelihood estimate (MLE) objective. However, in our VAE setting, the decoder conditions on the vector $z$ dynamically drew from a Gaussian distribution, instead of purely from previously generated tokens. Specifically, our decoder generates auto-regressively via:

\begin{equation}
p_{\boldsymbol{\theta}}(\boldsymbol{x} \mid \boldsymbol{z})=\prod_{t=1}^T p_{\boldsymbol{\theta}}\left(x_t \mid x_{<t}, \boldsymbol{z}\right).
\end{equation}

As stated above in Part \ref{vaeEncoder}, the intractable posterior for $z_n$ is approximated by $q_\phi\left(z_n \mid \boldsymbol{x}_{\leq n}, \boldsymbol{z}_{<n}\right)$ in Equation \ref{eq:encoder_formula}. Now we see the difference in the VAE decoder: the generation relies on high-level semantics and has the ability to produce a compact representation. 

\subsubsection{ELBO Training}
Both the encoder and decoder need to be trained to optimize their parameters. Supported by the above approximation, the entire training objective can be interpreted as evidence lower bound objective (ELBO): 

\begin{equation*}
\begin{aligned}
& \log p_{\boldsymbol{\theta}}(\boldsymbol{x}) \geq \mathcal{L}_{\text {ELBO }}= \\
& \mathbb{E}_{q_\phi(\boldsymbol{z} \mid \boldsymbol{x})}\left[\log p_{\boldsymbol{\theta}}(\boldsymbol{x} \mid \boldsymbol{z})\right] -\operatorname{KL}\left(q_{\boldsymbol{\phi}}(\boldsymbol{z} \mid \boldsymbol{x}) \| p(\boldsymbol{z})\right). 
\end{aligned}
\end{equation*}

Detailed derivations are emitted for clear idea depiction. In practice, we apply the re-parametrization trick~\cite{DBLP:journals/corr/KingmaW13} to allow back-propagation through all deterministic nodes and for efficient learning. 

Intuitively, we consider the first term as an auto-encoder objective, since it requires the model to do reconstruction based on the intermediate latent. The second term defines the KL divergence between the real distribution $q_{\boldsymbol{\phi}}(\boldsymbol{z} \mid \boldsymbol{x})$ and $p(\boldsymbol{z})$. To better implement these objectives, we refer to the regularized version of ELBO~\cite{li-etal-2020-optimus}, where ELBO is considered as the linear combination of reconstruction error and KL regularizer: 

\begin{equation}
\begin{aligned}
\mathcal{L}_\text{variational} & =\mathcal{L}_E+\beta \mathcal{L}_R, \text { with } \\
\mathcal{L}_E & =-\mathbb{E}_{q_\phi(\boldsymbol{z} \mid \boldsymbol{x})}\left[\log p_{\boldsymbol{\theta}}(\boldsymbol{x} \mid \boldsymbol{z})\right] \\
\mathcal{L}_R & =\operatorname{KL}\left(q_\phi(\boldsymbol{z} \mid \boldsymbol{x}) \| p(\boldsymbol{z})\right).
\end{aligned}
\end{equation}

Finally, we treat the loss term $ \mathcal{L}_\text{variational} $ as one of our minimization objectives in the following multi-task learning.

\begin{figure*}[ht]
	\begin{center}
		\includegraphics[width=\linewidth]{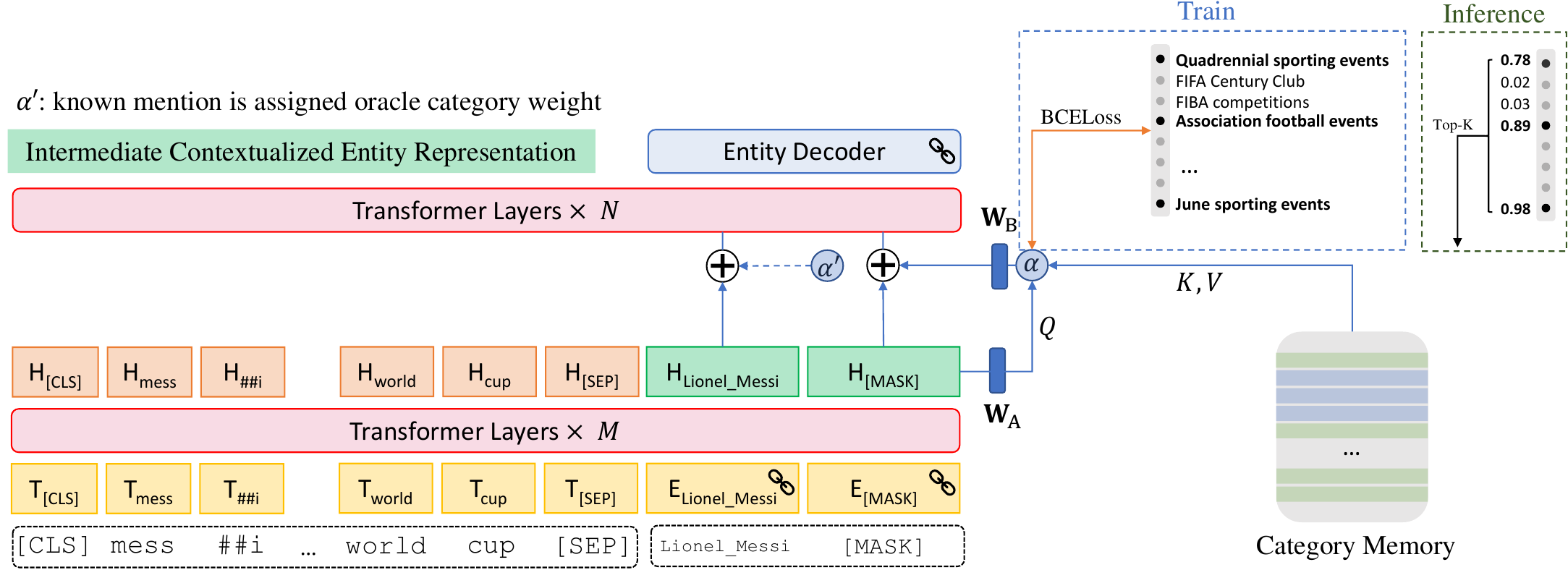}
	\end{center}
	\caption{Category memory illustration. The difference between training and inference is depicted using two dotted colored box.}
	\label{fig:category_memory}
\end{figure*}

\subsection{Category Memory}

We now formally define the category memory layer inserted into the intermediate Transformer layers. Let $\mathcal{E} = \{e_1, e_2, \ldots, e_n\}$ be the set of all possible candidate entities, and correspondingly, each entity $e_i$ has a predefined set of categories $C_{e_i}$. The union set of category sets of all candidate entities will be $C$. Based on this motivation, we construct a category vocabulary and a category embedding table $\mathbf{C} \in \mathbb{R}^{|C| \times d_\text{category}}$. The vocabulary establishes a mapping from textual category labels to indices.
As sometimes category labels can be too fine-grained, we refer readers to Appendix \ref{app:category} for the detailed design of category system.
% \footnote{E.g., \{\text{"June sporting events":} 14234, \text{"Quadrennial sporting events":} 48421, \ldots\} }

The embedding table $\mathbf{C}$ stores category representations that can be updated during massive pre-training. To be specific, we first formulate our model's forward (Figure~\ref{fig:category_memory}) as follows:

\begin{equation*}
    \begin{aligned}
    & \mathbf{T}^1, \mathbf{W}^1, \mathbf{E}^1=\text{Transformer}_{M}\left(\mathbf{T}^0, \mathbf{W}^0, \mathbf{E}^0\right), \\
    & \mathbf{H}=\text{CategoryMemory}\left(\mathbf{E}^1\right), \\
    & \mathbf{E}^{1 \prime}=\operatorname{LayerNorm}\left(\mathbf{H}+\mathbf{E}^1\right), \\
    & \mathbf{T}^2, \mathbf{W}^2, \mathbf{E}^2=\text{Transformer}_{N}\left(\mathbf{T}^1, \mathbf{W}^1, \mathbf{E}^{1 \prime}\right), 
    \end{aligned}
\end{equation*}
where symbols $\mathbf{T}$, $\mathbf{W}$, $\mathbf{E}$ stand for hidden states of topics, words and entities respectively. 

Each unresolved entity in the document is assigned with \texttt{[MASK]} token in the input sequence. During the forward pass, all intermediate entity representations corresponding to \texttt{[MASK]} token $e^i_{\text{masked}}$ are projected from $\mathbb{R}^{d_\text{entity}}$ to $\mathbb{R}^{d_\text{category}}$ using a linear layer without bias terms: 

\begin{equation}
    \mathbf{E}^i_{masked} = \mathbf{W}_{\mathbf{A}}\cdot e^i_{\text{masked}}
\end{equation}

Subsequently, the adapted intermediate entity representations will query all entries in the category embedding table. The aggregated weighted hidden states are computed as follows:

\begin{equation}
    \begin{aligned}
    \mathbf{H}_s^\text{category} &= \mathbf{W}_{\mathbf{B}}\left(\sum_{j=1}^{|C|} \alpha_{ij} \cdot \mathbf{C}^j\right), \\
    % \text{where} \ \alpha_j &= \operatorname{sigmoid}(\mathbf{C}^j \cdot (\mathbf{E}^i_{masked})^{T}).
    \end{aligned}
\end{equation}
where $\alpha_{ij} = \operatorname{sigmoid}(\mathbf{C}^j \cdot (\mathbf{E}^i_{\text{masked}})^{T})$ denotes matching score between $i$-th masked entity and $j$-th category. 
$\mathbf{W}_{\mathbf{B}}$ is a linear projection layer for dimension matching. 
During training,
% , since all category labels for masked entities are visible, 
we apply direct supervision from gold category guidance via binary cross-entropy loss:

% Zilin: check if binary CELoss can be interpreted as follows...
\begin{equation}
\begin{aligned}
\mathcal{L}_\text{category} & = - \frac{1}{|C|} \sum_{j=1}^{|C|} \alpha_{ij} \cdot \mathbb{I}_{\text{oracle}}, \\
% \text{where} \ \alpha_j & = \operatorname{sigmoid}(\mathbf{C}^j \cdot (\mathbf{E}^i_{masked})^{T}).
\end{aligned}
\end{equation}
where 
% $\alpha_j = \operatorname{sigmoid}(\mathbf{C}^j \cdot (\mathbf{E}^i_{\text{masked}})^{T})$
$\mathbb{I}_{\text{oracle}}$ denotes the indicator function of the oracle category labels for the $i$-th masked entity. 
% We use sigmoid activation instead of softmax since an entity usually involves multiple categories. 

\subsection{Multi-task Pre-training}
This part discusses the pre-training stage for \modelname.
% especially topic, category and disambiguation losses. 
First, we define the disambiguation loss, which is analogous to the well-known masked language modeling objective. In each training step, 30\% of the entity tokens are replaced with a special \texttt{[MASK]} token. We employ a linear decoder at the end of our model to reconstruct the masked tokens, as shown in Equation \ref{eq:lineardecoder}. 
% Notably, the parameters of $\mathbf{W}_\mathbf{D}$ are tied to those of the entity embedding layer to maximize training efficiency.

\begin{equation}
    \hat{\mathbf{E}} = \operatorname{softmax}(\mathbf{W}_\mathbf{D}\cdot \mathbf{E}^2_{masked} + \mathbf{b}_\mathbf{D}). \label{eq:lineardecoder}
\end{equation}

Equation \ref{eq:MEMLoss} represents the cross entropy loss over the entity vocabulary, where $\mathbb{I}_{e_k}$ denotes the indicator function of the $k$-th masked entity's ground-truth.

\begin{equation}
    \mathcal{L}_\text{disambiguation} = - \frac{1}{N_\text{masked}} \sum_{k=1}^{N_\text{masked}} \hat{\mathbf{E}} \cdot \mathbb{I}_{e_k}. \label{eq:MEMLoss}
\end{equation}

By incorporating all these losses, we derive our final multi-task learning objective being: 

\begin{equation}
    \mathcal{L} = \mathcal{L}_\text{disambiguation} + \alpha \mathcal{L}_\text{variational} + \gamma \mathcal{L}_\text{category}, 
\end{equation}
where coefficients $\alpha$ and $\gamma$ control relative importance of two auxiliary tasks.

\subsection{\modelname Inference}  % Mention oracle guidance, topk inference and candidate selection here.
Given a document $ X $ with $ N $ mentions, $ M = \{m_1, m_2, \ldots, m_{N}\} $, we now describe the coherent ED inference process. 

Considering a Transformer with an input length limit being $L$ tokens, we reserve $k$ tokens for topic latent, $n_e$ for shallow entity representation input, leaving the word input window being $ L - k - n_e $ tokens. Note that $n_e$ indicates the number of mentions in a certain sentence and varies among different training batches, so we set $n_e$ to the maximum number of mentions within the batch. We refer interested readers to Appendix \ref{app:prepare_tokens} for how we sample topic sentences and prepare input tokens.

With all input tokens ready, we predict entities for $N$ steps. Unlike a language generator which decodes the next token, at each step $i$, the model decodes all \texttt{[MASK]} tokens into entity predictions by selecting maximum indices in the logits. The entity prediction at step $i$ is decided using the highest-confidence strategy, \ie the entity with the highest log probability is resolved, while others have to wait until the next step. 

It is worth noting that utilizing candidate entity information can significantly reduce noisy predictions, as indicated by the green bars in Figure \ref{baseline_overview}. During inference in the category memory layer, only top-$k$ category entries are selected for weighted aggregation. 

% \begin{equation}
% \operatorname{Top}_K(\mathcal{C} \mid \mathbf{E}^i_{masked})=\underset{c \in \mathcal{C}}{\operatorname{Top}_K} \quad \mathbf{E}^i_{masked} \cdot c.
% \end{equation}

Once an entity prediction is determined, the category memory ceases to be queried at that position and instead receives a real category indicator to aggregate entries from the category memory. We refer to this as \textbf{oracle category guidance} because it allows for potential category-level guidance in disambiguating remaining mentions.

% Now with all input tokens ready, we delve into details of how to get ED predictions using such architecture. 

% Let $ X $ be a document with $ N $ mentions, $ M = \{m_1, m_2, \ldots, m_{|M|}\} $, where each of mentions $m_i$ is assigned to a certain number of entity candidates  $C_i = \{e_{i1}, e_{i2}, \ldots, e_{i|C_i|}\}$. Given a KB with a set of triplets $ G = \{(h, r, t)|h, t \in \mathcal{E}, r \in \mathcal{R}\} $, where $h$, $r$ and $t$ denote the head entity, relation and tail entity respectively, the goal of entity disambiguation is to link mentions $m_i$ to one of the corresponding mention candidates $C_i \subseteq \mathcal{E}$. 

\section{Experiments}

\begin{table*}[]
\centering
  \setlength{\tabcolsep}{2.6pt}
\begin{tabular}{lcccccc|cc}
\toprule[1pt]
Method                               & AIDA                 & MSNBC         & AQUAINT       & ACE2004       & CWEB          & WIKI          & AVG-6  &  AVG-5         \\ \midrule
\citet{DBLP:journals/pvldb/YosefHBSW11}                                 & 78.0                 & 79.0          & 56.0          & 80.0          & 58.6          & 63.0          & 69.1    & 67.3     \\
\citet{DBLP:conf/sigir/HulstHDBV20}                                  & 89.4                 & 90.7          & 84.1          & 85.3          & 71.9          & 73.1          & 82.4      & 81.0   \\
\citet{decao2021}                                & \textbf{93.3}$^\dag$ & 94.3          & 89.9          & 90.1          & 77.3          & 87.4          & $ 88.7^\dag$ & 87.8 \\
\citet{DBLP:conf/cidr/OrrLG0ALR21}$^*$                & 80.9                 & 80.5          & 74.2          & 83.6          & 70.2          & 76.2          & 77.6     &  76.9     \\
\citet{yang-etal-2018-collective}                                & \uline{93.0}$^\dag$  & 92.6          & 89.9          & 88.5          & \textbf{81.8} & 79.2          & $ 87.5^\dag$  &  86.4 \\
\citet{barba-etal-2022-extend}                               & 92.6$^\dag$          & 94.7          & 91.6          & 91.8          & 77.7          & 88.8          & 89.5$^\dag$ 
 &  88.9 \\
\citet{ayoola-etal-2022-refined}                              & 87.5                 & 94.4          & 91.8          & 91.6          & 77.8          & 88.7          & 88.6    & 88.8      \\
% \citet{DBLP:conf/naacl/AyoolaFP22}                               & 90.4                 & 94.8          & 92.6          & \textbf{93.4} & 78.2          & 90.4          & 90.0   & 89.9       \\
% $\text{GlobalED}_\text{base}^{*}$~\shortcite{yamada-etal-2022-global}  & -                    & 94.3          & 93.4          & {\ul 92.3}    & 76.7          & 87.6          & - & 88.8          \\
\citet{yamada-etal-2022-global}       & -                    & \textbf{96.3} & 93.5          & 91.9          & 78.9          & 89.1          & - & 89.9          \\ 
\midrule 
% \\[-1em]
Our $\text{CoherentED}_\text{base}$  & 88.2                 & 94.9          & 93.7          & {\ul 92.3}    & 77.2          & 87.8          & 89.0   &  89.2     \\
Our $\text{CoherentED}_\text{large}$ & 89.4                 & \textbf{96.3} & \textbf{94.6} & \textbf{93.4} & {\ul 81.1}    & \textbf{90.6}    & \textbf{90.9}  & \textbf{91.2} \\ 

\midrule 

\multicolumn{8}{c}{Model Ablations on $\text{CoherentED}_\text{large}$}                                                                                     \\ \midrule 
\quad \small{- w/o Topic Tokens}                     & 88.4                 & 94.6          & 93.4          & {\ul 92.3}    & 77.9          & 88.9          & 89.3   & 89.4       \\
\quad \small{- w/o Category Memory}                  & 89.1                 & {\ul 95.6}    & {\ul 93.8}    & {\ul 92.3}    & 79.8          & 90.1          & 90.1  & {\ul 90.3}  \\
\quad \small{- w/o Category Oracle Guidance}         & 89.9                 & 95.2          & 92.1          & \textbf{93.4} & 80.2          & \textbf{90.8} & {\ul 90.3}    &   {\ul 90.3}  \\

\bottomrule[1pt]
\end{tabular}
\caption{ED InKB micro F1 scores on test datasets. The best value is in \textbf{bold} and the second best is in \uline{underline}. $^*$ means results come from reproduced results of official open-source code. $^\dag$ indicates non-comparable metrics due to an unfair experimental setting. - indicates not reported in the original paper. For direct comparison with \citet{yamada-etal-2022-global}, AVG-5 reports average micro F1 scores on all test datasets except AIDA.}
\label{tab:peer}
\end{table*}

\subsection{Datasets and Settings \label{dataset_setting}}
For a fair comparison with previous works, we adopt the exact same settings used by \citet{decao2021}. Specifically, we borrow pre-generated candidate entity sets from \cite{le-titov-2018-improving}. Only entities with the top 30 $\hat{p}(e \mid m)$ score are considered candidates, and failure to include the oracle answer in the candidate set leads to a false negative prediction. For evaluation metrics, we report \emph{InKB} micro F1 scores on test splits of AIDA-CoNLL dataset~\cite{hoffart-etal-2011-robust} (AIDA), cleaned version of MSNBC, AQUAINT, ACE2004, WNED-CWEB (CWEB) and WNED-WIKI (WIKI)~\cite{DBLP:journals/semweb/GuoB18}. For training data, we use the 2022-09-01 Wikipedia dump without any weak or pseudo labels utilized in \citet{DBLP:conf/cidr/OrrLG0ALR21, broscheit-2019-investigating}. We do not run hyperparameter search due to limited training resources. Note that a few previous works use a mixture of AIDA and Wikipedia as training split, which we will indicate in the table caption as an unfair setting. Refer to Appendix~\ref{app:data} for dataset details, and Appendix~\ref{app:imple} for implementation, training and hyperparameter choice.

% model parameter comparison (与GlobalED公平的对比：VAE(3+3) + 6 -> globaled_base; VAE(6+6) + 12 -> globalED_large)
Two variants are proposed for a fair comparison with baseline models in terms of the number of Transformer layers. $\modelname_\text{base}$ contains 3, 3, and 6 layers for the VAE encoder, decoder and base model respectively, while in $\modelname_\text{large}$ these numbers become 6, 6 and 12. Both variants are with a comparable number of parameters with \citet{yamada-etal-2022-global}, namely 210M for the base variant and 440M for the large one. 

% with NVLink enabled. Gradient accumulation is enabled to expand the effective batch size.

% \usepackage[normalem]{ulem}

\subsection{Main Results}
We report peer comparisons in Table \ref{tab:peer}. Note that we consider the Wikipedia-only training setting, meaning no further fine-tuning or mixture training on AIDA is allowed and all evaluations are out-of-domain (OOD) tests. 
% For those peers who do not report their OOD results on AIDA, we denote them with $^{**}$, meaning such numbers are not for direct comparison. 
% Refer to Appendix \ref{app:peers} for brief introductions of selected peers.

In general, we achieve new state-of-the-art results on all test datasets except CWEB and AIDA datasets, surpassing the previous best by 1.3 F1 points and eliminating 9\% errors. On the CWEB dataset, our work still shows superiority over other neural-based methods, as the lengthy samples are too unfriendly to be understood globally by native neural encoders. On the ACE2004 dataset, since the number of mentions is relatively small, many reported numbers are identical. On other datasets, the relative improvements are consistent for $\modelname_\text{large}$ even though only additional cheap category labels are provided during pre-training. Such improvements also confirm the outstanding OOD ability of our methods since no fine-tuning is conducted on the downstream test datasets.

\subsection{Ablation Study}

% \PLACEHOLDER{05-18 TODO: need a separate ablation table}
In the lower part of Table \ref{tab:peer}, we report ablation experiments of proposed methods. All ablations are conducted on $\modelname_\text{large}$. 

Compared with the model without topic token injections, $\modelname_\text{large}$ improves greatly from 89.3 to 90.9 in average micro F1 score with a particular gain on the lengthy CWEB dataset from 77.9 to 81.1. Such performance gain on CWEB only falls short of \citet{yang-etal-2018-collective} which requires extensive feature engineering targeted at document-level representations. All other test sets also benefit from the powerful abstract modeling ability of topic VAE.

Compared with the model without a category memory layer, the $\text{CoherentED}_\text{large}$ improves from 90.1 to 90.9, not as significant as the improvement of topic injections but still notable enough. Note that category memory does not bring consistent performance gain on all test datasets, possibly because not all test samples are sensitive to category-level coherence. Furthermore, we attempt to disable the category oracle guidance strategy during evaluation, meaning for predicted entities in step-by-step ED, we still query the external category memory and aggregate the retrieved entries. 
And ablation shows that the oracle guidance does have a positive impact on the overall performance metrics.

\begin{figure}[ht]
	\begin{center}
		\includegraphics[width=\linewidth]{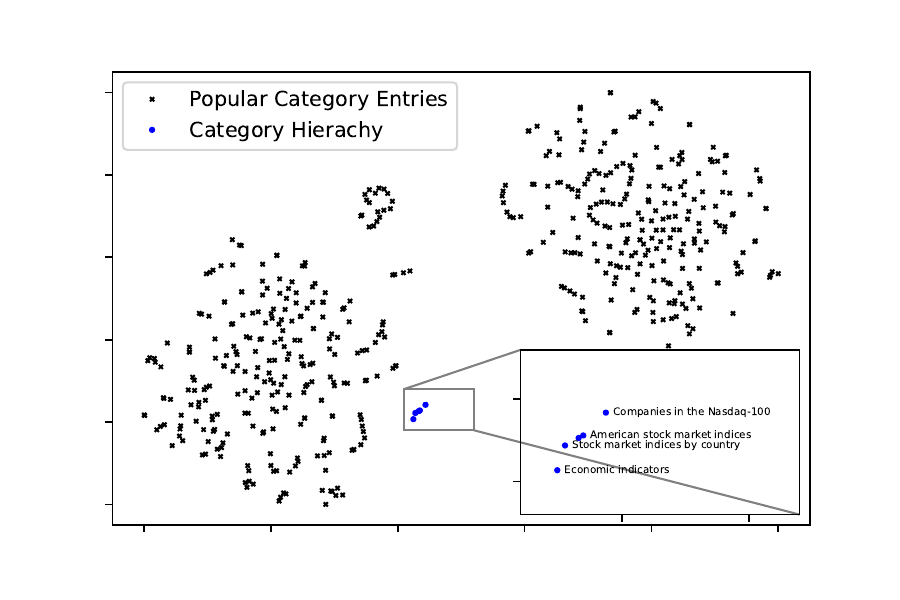}
	\end{center}
	\caption{T-SNE visualization of category embeddings after pre-training. Zoomed part denotes the specific category labels and corresponding sample points. Category "Stock market indices" is emitted for clear depiction.}
	\label{fig:vis_category}
\end{figure}

\subsection{Case Analysis and Visualization}

Besides ablations on evaluation metrics, we conduct a deeper analysis of proposed methods by visualizing data samples. Specifically, we perform t-SNE on category memory entries after joint training and topic vectors of sentences in MSNBC dataset.

% put case study and visualization here TBD
% case study 1: wrong in original global ED in MSNBC -> correct in CoherentED, due to topic modeling
% case study 2: wrong in original global ED in CWEB -> correct in CoherentED, due to topic modeling for lengthy documents
% case study 3: wrong in original global ED in some datasets -> correct in CoherentED, due to category dependency

% t-SNE visualization 1: category entries grouped by hierarchy
To validate the effectiveness of the learned category memory layer, we expect the embeddings of category entries stored in the memory to exhibit a certain degree of similarity with the category hierarchy in Wikipedia, \ie similar category entries are close to each other. In Figure \ref{fig:vis_category}, we present t-SNE visualization of the top 500 popular category entries and additional colored category group, where black cross data points represent popular category entries and blue circular points represent examples of a structured hierarchy from the Wikipedia category system\footnote{Economic indicators $\rightarrow$ Stock market indices $\rightarrow$ American stock market indices $\rightarrow$ Companies in the Nasdaq-100}.

% t-SNE visualization 2: topic latent grouped by topics in MSNBC & grouped by https://www.enchantedlearning.com/wordlist/ (refer to ToDKaT)

\begin{figure}[ht]
	\begin{center}
		\includegraphics[width=\linewidth]{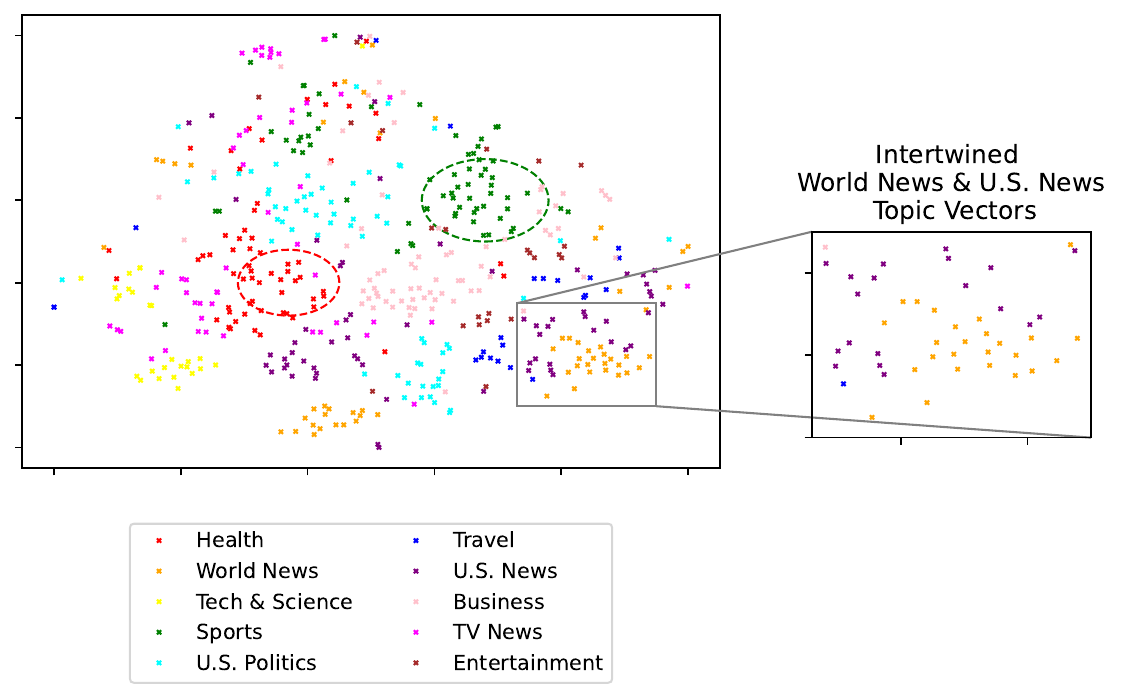}
	\end{center}
	\caption{T-SNE visualization of sentence vectors extracted by the topic probe strategy. Some polarized topic groups such as "Health" and "Sports" are denoted using colored dashed circles. Zoomed part reveals the intertwined nature between similar topics (\eg "World News" and "U.S. News"). Best viewed in color.}
	\label{fig:vis_topic}
\end{figure}

To evaluate the topic representation ability of our jointly trained topic VAE, we design an elegant probe strategy to investigate the topic modeling ability of \modelname. 
% As stated in Part \ref{dataset_setting},
MSNBC covers 20 documents on 10 topics\footnote{Business, U.S. Politics, Entertainment, Health, Sports, Tech \& Science, Travel, TV News, U.S. News, and World News.}.
By feeding each sentence in the MSNBC test set along with predicted entities tokens, we extract the \texttt{[CLS]} representation of $\text{CoherentED}_\text{large}$ and run t-SNE on these joint representations.

In Figure \ref{fig:vis_topic}, topic latent vectors of sentences in these documents are plotted into 511 data points, whose colors denote their oracle topic labels. We see that the majority of sentences under the same topic cluster into polarized groups, despite a few outliers possibly because they are for general purposes such as greeting and describing facts. Consistent with our expectations, similar topics are intertwined as they share high-level semantics to some extent. 

\section{Conclusion}
% limitation discussed: scalability & efficiency
We propose a novel entity disambiguation method \modelname, which injects latent topic vectors and utilizes cheap category knowledge sources to produce coherent disambiguation predictions. 
Specifically, we introduce an unsupervised topic variational auto-encoder and an external category memory bank to mitigate inconsistent entity predictions. Experimental results demonstrate the effectiveness of our proposed methods in terms of accuracy and coherence. Analysis of randomly picked cases and vector visualizations further confirm such effectiveness.

\section*{Limitations}
Still, our \modelname remains with two limitations: scalability and performance. Future works are expected to alleviate these limitations. First, \modelname can hardly handle emerging entities as this requires extending both the entity embedding layer and category memory layer. The evaluation metrics will degrade if no further training is conducted after such expansion. Second, despite the count of parameters and FLOPs of \modelname being quite comparable with baseline models, the advantage of coherent prediction only reveals itself in the scenario of step-by-step reasoning, \ie mentions are resolved one by one. This means multiple forward passes are needed for each document to achieve the most accurate results.

% Ethics Part is not mandatory
% \section*{Ethics Statement}

% Scientific work published at EMNLP 2023 must comply with the \href{https://www.aclweb.org/portal/content/acl-code-ethics}{ACL Ethics Policy}. We encourage all authors to include an explicit ethics statement on the broader impact of the work, or other ethical considerations after the conclusion but before the references. The ethics statement will not count toward the page limit (8 pages for long, 4 pages for short papers).

% \section*{Acknowledgements}
% Placeholder

\bibliography{anthology,custom}

\begin{thebibliography}{35}
\expandafter\ifx\csname natexlab\endcsname\relax\def\natexlab#1{#1}\fi

\bibitem[{Ayoola et~al.(2022)Ayoola, Tyagi, Fisher, Christodoulopoulos, and Pierleoni}]{ayoola-etal-2022-refined}
Tom Ayoola, Shubhi Tyagi, Joseph Fisher, Christos Christodoulopoulos, and Andrea Pierleoni. 2022.
\newblock \href {https://doi.org/10.18653/v1/2022.naacl-industry.24} {{R}e{F}in{ED}: An efficient zero-shot-capable approach to end-to-end entity linking}.
\newblock In \emph{Proceedings of the 2022 Conference of the North American Chapter of the Association for Computational Linguistics: Human Language Technologies: Industry Track}, pages 209--220, Hybrid: Seattle, Washington + Online. Association for Computational Linguistics.

\bibitem[{Barba et~al.(2022)Barba, Procopio, and Navigli}]{barba-etal-2022-extend}
Edoardo Barba, Luigi Procopio, and Roberto Navigli. 2022.
\newblock \href {https://doi.org/10.18653/v1/2022.acl-long.177} {{E}xt{E}n{D}: Extractive entity disambiguation}.
\newblock In \emph{Proceedings of the 60th Annual Meeting of the Association for Computational Linguistics (Volume 1: Long Papers)}, pages 2478--2488, Dublin, Ireland. Association for Computational Linguistics.

\bibitem[{Beltagy et~al.(2020)Beltagy, Peters, and Cohan}]{DBLP:journals/corr/abs-2004-05150}
Iz~Beltagy, Matthew~E. Peters, and Arman Cohan. 2020.
\newblock \href {http://arxiv.org/abs/2004.05150} {Longformer: The long-document transformer}.
\newblock \emph{CoRR}, abs/2004.05150.

\bibitem[{Blanco et~al.(2015)Blanco, Ottaviano, and Meij}]{DBLP:conf/wsdm/BlancoOM15}
Roi Blanco, Giuseppe Ottaviano, and Edgar Meij. 2015.
\newblock \href {https://doi.org/10.1145/2684822.2685317} {Fast and space-efficient entity linking for queries}.
\newblock In \emph{Proceedings of the Eighth {ACM} International Conference on Web Search and Data Mining, {WSDM} 2015, Shanghai, China, February 2-6, 2015}, pages 179--188. {ACM}.

\bibitem[{Bowman et~al.(2016)Bowman, Vilnis, Vinyals, Dai, Jozefowicz, and Bengio}]{bowman-etal-2016-generating}
Samuel~R. Bowman, Luke Vilnis, Oriol Vinyals, Andrew Dai, Rafal Jozefowicz, and Samy Bengio. 2016.
\newblock \href {https://doi.org/10.18653/v1/K16-1002} {Generating sentences from a continuous space}.
\newblock In \emph{Proceedings of the 20th {SIGNLL} Conference on Computational Natural Language Learning}, pages 10--21, Berlin, Germany. Association for Computational Linguistics.

\bibitem[{Broscheit(2019)}]{broscheit-2019-investigating}
Samuel Broscheit. 2019.
\newblock \href {https://doi.org/10.18653/v1/K19-1063} {Investigating entity knowledge in {BERT} with simple neural end-to-end entity linking}.
\newblock In \emph{Proceedings of the 23rd Conference on Computational Natural Language Learning (CoNLL)}, pages 677--685, Hong Kong, China. Association for Computational Linguistics.

\bibitem[{Bunescu and Pa{\c{s}}ca(2006)}]{bunescu-pasca-2006-using}
Razvan Bunescu and Marius Pa{\c{s}}ca. 2006.
\newblock \href {https://aclanthology.org/E06-1002} {Using encyclopedic knowledge for named entity disambiguation}.
\newblock In \emph{11th Conference of the {E}uropean Chapter of the Association for Computational Linguistics}, pages 9--16, Trento, Italy. Association for Computational Linguistics.

\bibitem[{Cao et~al.(2021)Cao, Izacard, Riedel, and Petroni}]{decao2021}
Nicola~De Cao, Gautier Izacard, Sebastian Riedel, and Fabio Petroni. 2021.
\newblock \href {https://openreview.net/forum?id=5k8F6UU39V} {Autoregressive entity retrieval}.
\newblock In \emph{9th International Conference on Learning Representations, {ICLR} 2021, Virtual Event, Austria, May 3-7, 2021}. OpenReview.net.

\bibitem[{Chisholm and Hachey(2015)}]{chisholm-hachey-2015-entity}
Andrew Chisholm and Ben Hachey. 2015.
\newblock \href {https://doi.org/10.1162/tacl_a_00129} {Entity disambiguation with web links}.
\newblock \emph{Transactions of the Association for Computational Linguistics}, 3:145--156.

\bibitem[{Daiber et~al.(2013)Daiber, Jakob, Hokamp, and Mendes}]{DBLP:conf/i-semantics/DaiberJHM13}
Joachim Daiber, Max Jakob, Chris Hokamp, and Pablo~N. Mendes. 2013.
\newblock \href {https://doi.org/10.1145/2506182.2506198} {Improving efficiency and accuracy in multilingual entity extraction}.
\newblock In \emph{{I-SEMANTICS} 2013 - 9th International Conference on Semantic Systems, {I-SEMANTICS} '13, Graz, Austria, September 4-6, 2013}, pages 121--124. {ACM}.

\bibitem[{De~Cao et~al.(2022)De~Cao, Wu, Popat, Artetxe, Goyal, Plekhanov, Zettlemoyer, Cancedda, Riedel, and Petroni}]{decao2022}
Nicola De~Cao, Ledell Wu, Kashyap Popat, Mikel Artetxe, Naman Goyal, Mikhail Plekhanov, Luke Zettlemoyer, Nicola Cancedda, Sebastian Riedel, and Fabio Petroni. 2022.
\newblock Multilingual autoregressive entity linking.
\newblock \emph{Transactions of the Association for Computational Linguistics}, 10:274--290.

\bibitem[{Devlin et~al.(2019)Devlin, Chang, Lee, and Toutanova}]{devlin-etal-2019-bert}
Jacob Devlin, Ming-Wei Chang, Kenton Lee, and Kristina Toutanova. 2019.
\newblock \href {https://doi.org/10.18653/v1/N19-1423} {{BERT}: Pre-training of deep bidirectional transformers for language understanding}.
\newblock In \emph{Proceedings of the 2019 Conference of the North {A}merican Chapter of the Association for Computational Linguistics: Human Language Technologies, Volume 1 (Long and Short Papers)}, pages 4171--4186, Minneapolis, Minnesota. Association for Computational Linguistics.

\bibitem[{Doddington et~al.(2004)Doddington, Mitchell, Przybocki, Ramshaw, Strassel, and Weischedel}]{doddington-etal-2004-automatic}
George Doddington, Alexis Mitchell, Mark Przybocki, Lance Ramshaw, Stephanie Strassel, and Ralph Weischedel. 2004.
\newblock \href {http://www.lrec-conf.org/proceedings/lrec2004/pdf/5.pdf} {The automatic content extraction ({ACE}) program {--} tasks, data, and evaluation}.
\newblock In \emph{Proceedings of the Fourth International Conference on Language Resources and Evaluation ({LREC}{'}04)}, Lisbon, Portugal. European Language Resources Association (ELRA).

\bibitem[{Gabrilovich et~al.(2013)Gabrilovich, Ringgaard, and Subramanya}]{FACC1}
Evgeniy Gabrilovich, Michael Ringgaard, and Amarnag Subramanya. 2013.
\newblock Facc1: Freebase annotation of clueweb corpora, version 1 (release date 2013-06-26, format version 1, correction level 0).

\bibitem[{Guo and Barbosa(2018)}]{DBLP:journals/semweb/GuoB18}
Zhaochen Guo and Denilson Barbosa. 2018.
\newblock \href {https://doi.org/10.3233/SW-170273} {Robust named entity disambiguation with random walks}.
\newblock \emph{Semantic Web}, 9(4):459--479.

\bibitem[{Hoffart et~al.(2011)Hoffart, Yosef, Bordino, F{\"u}rstenau, Pinkal, Spaniol, Taneva, Thater, and Weikum}]{hoffart-etal-2011-robust}
Johannes Hoffart, Mohamed~Amir Yosef, Ilaria Bordino, Hagen F{\"u}rstenau, Manfred Pinkal, Marc Spaniol, Bilyana Taneva, Stefan Thater, and Gerhard Weikum. 2011.
\newblock \href {https://aclanthology.org/D11-1072} {Robust disambiguation of named entities in text}.
\newblock In \emph{Proceedings of the 2011 Conference on Empirical Methods in Natural Language Processing}, pages 782--792, Edinburgh, Scotland, UK. Association for Computational Linguistics.

\bibitem[{Hoffmann et~al.(2011)Hoffmann, Zhang, Ling, Zettlemoyer, and Weld}]{hoffmann-etal-2011-knowledge}
Raphael Hoffmann, Congle Zhang, Xiao Ling, Luke Zettlemoyer, and Daniel~S. Weld. 2011.
\newblock \href {https://aclanthology.org/P11-1055} {Knowledge-based weak supervision for information extraction of overlapping relations}.
\newblock In \emph{Proceedings of the 49th Annual Meeting of the Association for Computational Linguistics: Human Language Technologies}, pages 541--550, Portland, Oregon, USA. Association for Computational Linguistics.

\bibitem[{Kingma and Welling(2014)}]{DBLP:journals/corr/KingmaW13}
Diederik~P. Kingma and Max Welling. 2014.
\newblock \href {http://arxiv.org/abs/1312.6114} {Auto-encoding variational bayes}.
\newblock In \emph{2nd International Conference on Learning Representations, {ICLR} 2014, Banff, AB, Canada, April 14-16, 2014, Conference Track Proceedings}.

\bibitem[{Le and Titov(2018)}]{le-titov-2018-improving}
Phong Le and Ivan Titov. 2018.
\newblock \href {https://doi.org/10.18653/v1/P18-1148} {Improving entity linking by modeling latent relations between mentions}.
\newblock In \emph{Proceedings of the 56th Annual Meeting of the Association for Computational Linguistics (Volume 1: Long Papers)}, pages 1595--1604, Melbourne, Australia. Association for Computational Linguistics.

\bibitem[{Li et~al.(2020)Li, Gao, Li, Peng, Li, Zhang, and Gao}]{li-etal-2020-optimus}
Chunyuan Li, Xiang Gao, Yuan Li, Baolin Peng, Xiujun Li, Yizhe Zhang, and Jianfeng Gao. 2020.
\newblock \href {https://doi.org/10.18653/v1/2020.emnlp-main.378} {Optimus: Organizing sentences via pre-trained modeling of a latent space}.
\newblock In \emph{Proceedings of the 2020 Conference on Empirical Methods in Natural Language Processing (EMNLP)}, pages 4678--4699, Online. Association for Computational Linguistics.

\bibitem[{Li et~al.(2022)Li, Li, Li, Li, Liu, Liu, and Dong}]{li2022improving}
Qijia Li, Feng Li, Shuchao Li, Xiaoyu Li, Kang Liu, Qing Liu, and Pengcheng Dong. 2022.
\newblock Improving entity linking by introducing knowledge graph structure information.
\newblock \emph{Applied Sciences}, 12(5):2702.

\bibitem[{Onoe and Durrett(2020)}]{DBLP:conf/aaai/OnoeD20}
Yasumasa Onoe and Greg Durrett. 2020.
\newblock \href {https://ojs.aaai.org/index.php/AAAI/article/view/6380} {Fine-grained entity typing for domain independent entity linking}.
\newblock In \emph{The Thirty-Fourth {AAAI} Conference on Artificial Intelligence, {AAAI} 2020, The Thirty-Second Innovative Applications of Artificial Intelligence Conference, {IAAI} 2020, The Tenth {AAAI} Symposium on Educational Advances in Artificial Intelligence, {EAAI} 2020, New York, NY, USA, February 7-12, 2020}, pages 8576--8583. {AAAI} Press.

\bibitem[{Orr et~al.(2021)Orr, Leszczynski, Guha, Wu, Arora, Ling, and R{\'{e}}}]{DBLP:conf/cidr/OrrLG0ALR21}
Laurel~J. Orr, Megan Leszczynski, Neel Guha, Sen Wu, Simran Arora, Xiao Ling, and Christopher R{\'{e}}. 2021.
\newblock \href {http://cidrdb.org/cidr2021/papers/cidr2021\_paper13.pdf} {Bootleg: Chasing the tail with self-supervised named entity disambiguation}.
\newblock In \emph{11th Conference on Innovative Data Systems Research, {CIDR} 2021, Virtual Event, January 11-15, 2021, Online Proceedings}. www.cidrdb.org.

\bibitem[{Radford et~al.(2019)Radford, Wu, Child, Luan, Amodei, Sutskever et~al.}]{radford2019language}
Alec Radford, Jeffrey Wu, Rewon Child, David Luan, Dario Amodei, Ilya Sutskever, et~al. 2019.
\newblock Language models are unsupervised multitask learners.
\newblock \emph{OpenAI blog}, 1(8):9.

\bibitem[{Rasley et~al.(2020)Rasley, Rajbhandari, Ruwase, and He}]{deepspeed}
Jeff Rasley, Samyam Rajbhandari, Olatunji Ruwase, and Yuxiong He. 2020.
\newblock \href {https://doi.org/10.1145/3394486.3406703} {Deepspeed: System optimizations enable training deep learning models with over 100 billion parameters}.
\newblock In \emph{Proceedings of the 26th ACM SIGKDD International Conference on Knowledge Discovery \&amp; Data Mining}, KDD '20, page 3505–3506, New York, NY, USA. Association for Computing Machinery.

\bibitem[{Snell et~al.(2022)Snell, Klein, and Zhong}]{DBLP:journals/corr/abs-2209-15189}
Charlie Snell, Dan Klein, and Ruiqi Zhong. 2022.
\newblock \href {https://doi.org/10.48550/arXiv.2209.15189} {Learning by distilling context}.
\newblock \emph{CoRR}, abs/2209.15189.

\bibitem[{van Hulst et~al.(2020)van Hulst, Hasibi, Dercksen, Balog, and de~Vries}]{DBLP:conf/sigir/HulstHDBV20}
Johannes~M. van Hulst, Faegheh Hasibi, Koen Dercksen, Krisztian Balog, and Arjen~P. de~Vries. 2020.
\newblock \href {https://doi.org/10.1145/3397271.3401416} {{REL:} an entity linker standing on the shoulders of giants}.
\newblock In \emph{Proceedings of the 43rd International {ACM} {SIGIR} conference on research and development in Information Retrieval, {SIGIR} 2020, Virtual Event, China, July 25-30, 2020}, pages 2197--2200. {ACM}.

\bibitem[{Vaswani et~al.(2017)Vaswani, Shazeer, Parmar, Uszkoreit, Jones, Gomez, Kaiser, and Polosukhin}]{DBLP:conf/nips/VaswaniSPUJGKP17}
Ashish Vaswani, Noam Shazeer, Niki Parmar, Jakob Uszkoreit, Llion Jones, Aidan~N. Gomez, Lukasz Kaiser, and Illia Polosukhin. 2017.
\newblock \href {https://proceedings.neurips.cc/paper/2017/hash/3f5ee243547dee91fbd053c1c4a845aa-Abstract.html} {Attention is all you need}.
\newblock In \emph{Advances in Neural Information Processing Systems 30: Annual Conference on Neural Information Processing Systems 2017, December 4-9, 2017, Long Beach, CA, {USA}}, pages 5998--6008.

\bibitem[{Wingate et~al.(2022)Wingate, Shoeybi, and Sorensen}]{wingate-etal-2022-prompt}
David Wingate, Mohammad Shoeybi, and Taylor Sorensen. 2022.
\newblock \href {https://aclanthology.org/2022.findings-emnlp.412} {Prompt compression and contrastive conditioning for controllability and toxicity reduction in language models}.
\newblock In \emph{Findings of the Association for Computational Linguistics: EMNLP 2022}, pages 5621--5634, Abu Dhabi, United Arab Emirates. Association for Computational Linguistics.

\bibitem[{Wolf et~al.(2020)Wolf, Debut, Sanh, Chaumond, Delangue, Moi, Cistac, Rault, Louf, Funtowicz, Davison, Shleifer, von Platen, Ma, Jernite, Plu, Xu, Le~Scao, Gugger, Drame, Lhoest, and Rush}]{wolf-etal-2020-transformers}
Thomas Wolf, Lysandre Debut, Victor Sanh, Julien Chaumond, Clement Delangue, Anthony Moi, Pierric Cistac, Tim Rault, Remi Louf, Morgan Funtowicz, Joe Davison, Sam Shleifer, Patrick von Platen, Clara Ma, Yacine Jernite, Julien Plu, Canwen Xu, Teven Le~Scao, Sylvain Gugger, Mariama Drame, Quentin Lhoest, and Alexander Rush. 2020.
\newblock \href {https://doi.org/10.18653/v1/2020.emnlp-demos.6} {Transformers: State-of-the-art natural language processing}.
\newblock In \emph{Proceedings of the 2020 Conference on Empirical Methods in Natural Language Processing: System Demonstrations}, pages 38--45, Online. Association for Computational Linguistics.

\bibitem[{Yamada et~al.(2020)Yamada, Asai, Shindo, Takeda, and Matsumoto}]{yamada-etal-2020-luke}
Ikuya Yamada, Akari Asai, Hiroyuki Shindo, Hideaki Takeda, and Yuji Matsumoto. 2020.
\newblock \href {https://doi.org/10.18653/v1/2020.emnlp-main.523} {{LUKE}: Deep contextualized entity representations with entity-aware self-attention}.
\newblock In \emph{Proceedings of the 2020 Conference on Empirical Methods in Natural Language Processing (EMNLP)}, pages 6442--6454, Online. Association for Computational Linguistics.

\bibitem[{Yamada et~al.(2022)Yamada, Washio, Shindo, and Matsumoto}]{yamada-etal-2022-global}
Ikuya Yamada, Koki Washio, Hiroyuki Shindo, and Yuji Matsumoto. 2022.
\newblock \href {https://doi.org/10.18653/v1/2022.naacl-main.238} {Global entity disambiguation with {BERT}}.
\newblock In \emph{Proceedings of the 2022 Conference of the North American Chapter of the Association for Computational Linguistics: Human Language Technologies}, pages 3264--3271, Seattle, United States. Association for Computational Linguistics.

\bibitem[{Yang et~al.(2018)Yang, Irsoy, and Rahman}]{yang-etal-2018-collective}
Yi~Yang, Ozan Irsoy, and Kazi~Shefaet Rahman. 2018.
\newblock \href {https://doi.org/10.18653/v1/N18-1071} {Collective entity disambiguation with structured gradient tree boosting}.
\newblock In \emph{Proceedings of the 2018 Conference of the North {A}merican Chapter of the Association for Computational Linguistics: Human Language Technologies, Volume 1 (Long Papers)}, pages 777--786, New Orleans, Louisiana. Association for Computational Linguistics.

\bibitem[{Yih et~al.(2015)Yih, Chang, He, and Gao}]{yih-etal-2015-semantic}
Wen-tau Yih, Ming-Wei Chang, Xiaodong He, and Jianfeng Gao. 2015.
\newblock \href {https://doi.org/10.3115/v1/P15-1128} {Semantic parsing via staged query graph generation: Question answering with knowledge base}.
\newblock In \emph{Proceedings of the 53rd Annual Meeting of the Association for Computational Linguistics and the 7th International Joint Conference on Natural Language Processing (Volume 1: Long Papers)}, pages 1321--1331, Beijing, China. Association for Computational Linguistics.

\bibitem[{Yosef et~al.(2011)Yosef, Hoffart, Bordino, Spaniol, and Weikum}]{DBLP:journals/pvldb/YosefHBSW11}
Mohamed~Amir Yosef, Johannes Hoffart, Ilaria Bordino, Marc Spaniol, and Gerhard Weikum. 2011.
\newblock \href {http://www.vldb.org/pvldb/vol4/p1450-yosef.pdf} {{AIDA:} an online tool for accurate disambiguation of named entities in text and tables}.
\newblock \emph{Proc. {VLDB} Endow.}, 4(12):1450--1453.

\end{thebibliography}
\bibliographystyle{acl_natbib}

\appendix

\newpage

% \section{Inference Input Tokens Preparation}
% \label{app:trunc}
% % First, the document is tokenized into $L_D$ word tokens and split into sentences. If word tokens fit in the word input window (i.e. $L_D \leq L - k - n_e$), we utilize all word tokens and sample $k$ topic sentences uniformly. Otherwise, we truncate the document with the sentence to be disambiguated as the center, and prioritize sampling $k$ sentences outside the trimming range as the topic sentences. Different sampling strategies bring negligible improvement and will not be discussed here. Selected sentences get encoded through a pre-trained topic encoder and prepend their topic representations on the input sequence. Note that the topic decoder is no longer needed as it only supports variational learning in the auxiliary branch. Lastly, $N$ $[MASK]$ tokens are appended to the sequence, indicating all $N$ mentions are unresolved. For training samples where $N < n_e$, we concatenate more $N - n_e$ $[PAD]$ tokens.
\section{Category System Design}
\label{app:category}
Sometimes category labels can be too fine-grained, \eg \textit{Apple Inc.} is with one category label of \uline{Computer companies established in 1976}, which is less informative compared with two separate labels \uline{Computer companies} and \uline{in 1976}. To mitigate this issue, we extend category construction methods from \citet{DBLP:conf/aaai/OnoeD20}, where prepositions
\footnote{Specifically, prepositions refer to those that are frequently used in the category such as 'in', 'from', 'for', 'of', 'by', 'for', and 'involving'.} 
are considered as stop words to disassemble original category labels. We further ignore differences among prepositions, \ie \uline{of the United States} and \uline{in the United States} are considered to be indistinguishable and unified into \uline{[PERP] the United States}. This helps simplify the category labels and ensures that the model focuses on the relevant semantic information rather than specific prepositions.

\section{Evaluation Dataset Details}
\label{app:data}
The brief dataset descriptions are as follows:
\begin{enumerate}
    \item \textbf{AIDA} contains $18,448$ training samples, $4791$ validation samples and $4485$ test samples. It also served as one of the largest manually annotated EL and ED datasets. Note that other datasets contain test split only.
    \item \textbf{MSNBC} is a news corpus with 20 documents and 656 mentions. Despite its small scale, MSNBC covers 10 obvious topics and thus acts as a perfect testbed for our topic VAE approach.
    \item \textbf{AQUAINT} is another news corpus containing 50 documents from Xinhua News, the New York Times and the Associated Press, covering 727 samples.
    \item \textbf{ACE2004} is a manually annotated subset of \citet{doddington-etal-2004-automatic} containing 257 samples.
    \item \textbf{CWEB} is an automatically constructed dataset from ClueWeb corpus\footnote{\url{https://lemurproject.org/clueweb12}} in \citet{DBLP:journals/semweb/GuoB18}, containing 11,154 samples. It is worth mentioning that most ED works perform similarly on CWEB, partially because the average length of documents in CWEB is significantly longer than others. With approximately 1,700 words on average per document, none of the trivial BERT-based models can handle the entire document. \citet{yang-etal-2018-collective} got relatively better performance on it as their models involve heavy hand-crafted features designed to capture document-level semantics. We later show that our topic learning-based methods can achieve similar performance with minimal human involvement.  % Zilin: I don't remember which paper indicates this long-and-hard-to-capture situation...
    \item \textbf{WIKI} is another automatically extracted test dataset from \citet{FACC1}, covering 6,821 samples.
\end{enumerate}

\section{Preparing Input Tokens for \modelname}
\label{app:prepare_tokens}
First, the document is tokenized into $L_D$ text tokens and split into sentences. If text tokens fit in the word input window (i.e. $L_D \leq L - k - n_e$), we utilize all word tokens and sample $k$ topic sentences uniformly. Otherwise, we truncate the document with the sentence to be disambiguated as the center\footnote{Different sampling strategies bring negligible improvement. As such, we will not discuss performance differences caused by sampling.}, and prioritize sampling $k$ sentences outside the trimming range as the topic sentences. Selected sentences get encoded through a pre-trained topic encoder and prepend their topic representations on the input sequence. Note that the topic decoder is no longer needed as it only supports variational learning in the auxiliary branch. Lastly, $N$ \texttt{[MASK]} tokens are appended to the sequence, indicating all $N$ mentions are unresolved. For training samples where $N < n_e$, we concatenate more $N - n_e$ \texttt{[PAD]} tokens.

\section{Brief Introductions of Peers}
\label{app:peers}

\begin{itemize}
    \item \textbf{AIDA}~\cite{DBLP:journals/pvldb/YosefHBSW11} is a traditional framework and online tool for entity detection and disambiguation.
    \item \textbf{REL}~\cite{DBLP:conf/sigir/HulstHDBV20} is a modern open-source toolkit for entity linking equipped with customized deep models.
    \item \textbf{GENRE}~\cite{decao2021} is the first to formulate entity linking and disambiguation into the constrained text generation task via pre-defined trie. Auto-regressive decoding nature makes it hard for real-time usage.
    \item \textbf{Bootleg}~\cite{DBLP:conf/cidr/OrrLG0ALR21} focuses on modeling reasoning patterns for disambiguation in a self-supervised manner. Tail entities who rarely appear in KB and documents are especially investigated in this work.
    \item \textbf{BiBSG}~\cite{yang-etal-2018-collective} is the first to introduce the structured gradient tree boosting (SGTB) algorithm to collective entity disambiguation with many efforts in making use of global information from both the past and future to perform a better local search.
    \item \textbf{ExtEND}~\cite{barba-etal-2022-extend} formulates the ED problem into a span extraction task supported by a Longformer model that predicts entity span in the input sequence.
    \item \textbf{ReFinED}~\cite{ayoola-etal-2022-refined} is an efficient zero-shot end-to-end entity linker using score-based bi-encoder architecture, which seeks a trade-off between performance and efficiency.
    % \item \textbf{IEDRKB}~\cite{DBLP:conf/naacl/AyoolaFP22} is an extension of ReFinED but with more expertise in utilizing KB facts for ED, and it is also the previous SOTA in terms of the average micro F1 score.
    \item \textbf{GlobalED}~\cite{yamada-etal-2022-global} considers ED as a masked token prediction problem and is also the baseline of our work.
\end{itemize}

\begin{figure*}[ht]
	\begin{center}
		\includegraphics[width=\linewidth]{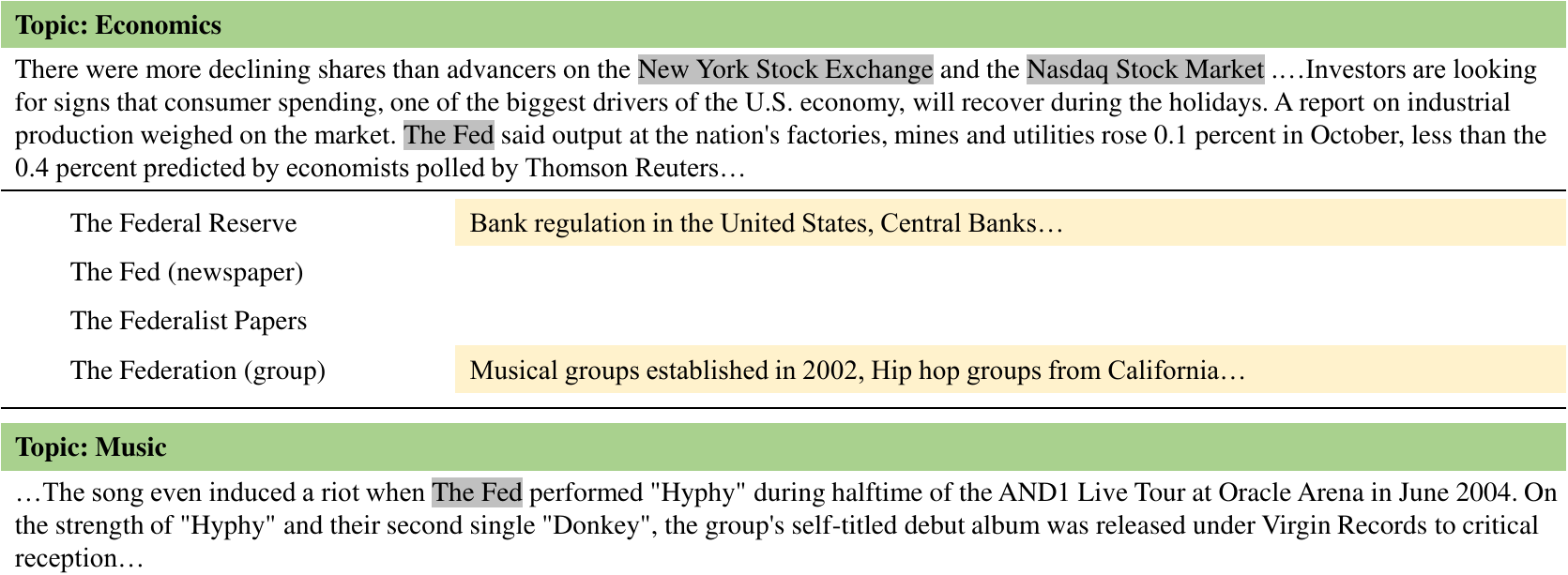}
	\end{center}
	\caption{ED case analysis revealing two critical motivations of our work, topic coherence and categorical coherence. Mentions are annotated with a gray background. Two documents share the same mention "The Fed". The document in the upper part is centered around the economics topic while the lower one elaborates on the music topic. The middle part lists four entity candidates for the mention "The Fed", with some corresponding category labels on yellow background.} 
	\label{fig:case}
\end{figure*}

\section{Implementation, Training and Hyperparameters}
\label{app:imple}

\begin{table}[ht]
\centering
\begin{tabular}{cc}
\toprule[1pt]
\textbf{Hyperparameter} & \textbf{Value} \\ \hline
learning rate (stage 1) & 5e-4           \\
learning rate (stage 2) & 5e-5           \\
weight\_decay           & 1e-2           \\
batch size per device   & 4              \\
effective batch size    & 2048           \\
learning rate strategy  & WarmupDecayLR  \\
optimizer               & AdamW          \\
dropout                 & 0.1       \\    
gradient clipping       & 1.0       \\    
\bottomrule[1pt]
\end{tabular}
\caption{Hyperparameters used for training \modelname}
\label{tab:hyperparameters}
\end{table}

We use the Huggingface~\cite{wolf-etal-2020-transformers} version of Transformer as the codebase. A total of $127,314$ entities are considered in entity vocabulary and entity prediction head, resulting in a category vocabulary of size $391,234$. DeepSpeed~\cite{deepspeed} is used for maximum hardware utilization and parallel training management. Table \ref{tab:hyperparameters} presents most of the hyperparameters in training. Due to the limited computation resource, we do not run massive hyperparameter searches. Multi-task coefficients are set to $\alpha = 0.1$ and $\gamma = 10$ with a few empirical trials done. During inference, the $k$ in Top-K category retrieval is set to 10, as the average number of categories of all entities present is close to this number.

We train our proposed model in two stages. In stage 1, we freeze all the parameters except for the fresh entity embedding layer and category memory layer. Consequently, the variational objective is disabled in stage 1. Then after 1 epoch, we activate all parameters and enable three objectives in stage 2, which lasts for 6 epochs. Pre-training a VAE can be difficult due to the notorious KL vanishing issue~\cite{bowman-etal-2016-generating}, causing the decoder completely ignores the topic latent $z$ in learning. As a practical solution to mitigate this, a cyclical schedule is applied to the KL regularizer coefficient $\beta$. 
The training takes approximately 1 day for $\modelname_\text{base}$ and 3 days $\modelname_\text{large}$ on 8 A100-SXM4-40GB GPUs.
% Refer to \cite{li-etal-2020-optimus} for a detailed annealing strategy. 

\section{Case Study}

In Figure~\ref{fig:case}, we illustrate document samples extracted from the MSNBC test dataset, wherein the mention "The Fed" can be readily disambiguated if provided with the corresponding topic.

Besides topic coherence, the relationship among entities also matters in the ED task. In the upper of Figure~\ref{fig:case}, mentions are highly correlated in their category level, and previous bi-encoder and cross-encoder solutions totally ignore the dependencies among entities and focus on learning representations alone. 

Consider the upper document in Figure~\ref{fig:case}, where three mentions are highlighted with a gray background. The correct linked entity for the mention "New York Stock Exchange" shares the exact category "Stock exchanges in the United States" with the correct linked entity for "Nasdaq Stock Market." Moreover, these two mentions can implicitly guide entity prediction for "The Fed" due to the high correlation between their respective categories.

\end{document}